\documentclass[10pt,twocolumn,letterpaper]{article}

\usepackage{cvpr}
\usepackage{times}
\usepackage{epsfig}
\usepackage{graphicx}
\usepackage{amsmath}
\usepackage{amssymb}
\usepackage{booktabs}
\usepackage{multirow}
\newcommand{\enum}[2]{\ensuremath{#1\text{e#2}}}

\begin{document}

%%%%%%%%% TITLE
\title{Face Re-morphing: Differential Morphing Attack Detection\\via Feature-Space Similarity Changes}

\author{Jie Jin$^{1}$  \quad Masakatsu Nishigaki$^{1}$ \quad Tetsushi Ohki$^{1,2}$\\
$^1$Graduate School of Science and Technology, Shizuoka University 
$^2$RIKEN AIP\\
3-5-1 Johoku, Chuo-ku, Hamamatsu, Shizuoka 432-8011\\
1-4-1 Nihonbashi, Chuo-ku, Tokyo 103-0027, Japan\\
{\tt\small jie.jin@sec.inf.shizuoka.ac.jp, \{nisigaki,ohki\}@inf.shizuoka.ac.jp}
}

\maketitle
\thispagestyle{empty}

%%%%%%%%% ABSTRACT
\begin{abstract}
Face morphing attacks pose a serious threat to face recognition systems because a single morphed document image can be matched to multiple contributors.
Differential morphing attack detection (D-MAD) addresses this threat by comparing a document image with a trusted live image, but existing methods often rely on static feature differences, constituent-face reconstruction, or multi-cue fusion.

This paper proposes Face Re-morphing, a D-MAD method that uses the feature-space response to an additional morphing operation as a detection cue.
Given a document image and a trusted live image, the proposed method generates a re-morphed image and uses the change between the document--live and live--re-morphed cosine similarities as the detection score.

Experiments on FRLL-Morphs and FEI Morph show that the proposed cue is effective across different morphing conditions, re-morphing methods, and face recognition models.
Comparisons with existing methods show favorable results on AMSL and indicate that the proposed method performs well under the Criminal condition on FEI Morph Version~1, particularly when using MorDIFF.
These results indicate that re-morphing-induced similarity change provides a complementary cue for D-MAD.
\end{abstract}

\section{Introduction}

Face recognition systems are widely used for person identification and identity verification, and have been deployed in many security systems and applications.
In particular, border control systems require high robustness and reliability, since authentication is performed by referring to facial images stored in passports or visas.
However, previous studies have shown that face recognition systems can be bypassed by using morphed face images created from two different individuals, making morphing attacks a serious threat to face recognition systems \cite{ferrara2014magic, damer2018morgan}.

In a morphing attack, an attacker presents a morphed image at enrollment.
This image is created by combining the facial images of the attacker and an accomplice.
If the morphed image is accepted as a legitimate document image, multiple individuals who contributed to the morph may be successfully matched against the same document image.
This means that a passport or identity document, which should correspond to only one person, can effectively be shared by multiple individuals, thereby undermining the one-to-one correspondence between a registered identity and a single subject that underlies border control and identity verification.
In particular, an accomplice with no criminal record may obtain a legitimate document using a morphed image, and a different person may later use that document.
This poses a serious risk to face-recognition-based operations, including automated border control.

Morphing Attack Detection (MAD) has been studied as a countermeasure against such attacks.
MAD can be broadly categorized into Single-image MAD (S-MAD), which determines whether a single document image is morphed, and Differential MAD (D-MAD), which utilizes both the document image and a live image captured in a trusted acquisition environment \cite{namis2024face}.
Although S-MAD can be applied when only a single image is available, detection may become difficult for high-quality morphed images with few visible artifacts.
In contrast, D-MAD can utilize both the document image and the live image, making it suitable for scenarios in which a trusted live image is available.

Existing D-MAD methods have evolved into approaches that directly exploit the difference between the document image and the live image, approaches that make decisions through the reconstruction of constituent faces contained in a morph image, and approaches that integrate multiple cues \cite{scherhag2020deep, ferrara2017face, borghi2021double}.
However, these methods often depend on how differences are modeled, on reconstruction quality, or on how multiple cues are integrated, which leaves challenges in achieving stable detection across different morphing conditions.

In this study, we propose Face Re-morphing, a D-MAD method that exploits changes in cosine similarity caused by applying an additional morphing operation between the document image and the live image.
Such changes can serve as a discriminative cue because their behavior is expected to differ depending on whether the document image is a bona-fide image or a morphed image.
For a bona-fide image, the document image is composed of the facial characteristics of a single subject, and thus the cosine similarity change induced by Re-morphing with the live image is expected to remain within a relatively natural range.
In contrast, when the document image is a morph, it already contains facial characteristics from multiple subjects, so the cosine similarity change produced by Re-morphing with the live image is expected to show a different tendency from that observed for bona-fide images.
The proposed method utilizes this difference for morphing attack detection.

The main contributions of this paper are twofold.
First, we introduce Face Re-morphing as a new detection concept for D-MAD, in which existing face morphing techniques are utilized as part of the detection process.
Second, using the cosine similarity change before and after re-morphing as the detection cue, we empirically demonstrate that the similarity change exhibits a consistent discriminative tendency across different morphing methods and feature extractors.

\section{Related Work}

\subsection{Feature Difference-Based D-MAD}
Feature difference-based D-MAD approaches identify morphing attacks by measuring differences between the document image and the live image.
These methods typically leverage cues including facial landmark misalignments, shape variations, deep embeddings, and identity matching scores.

As a representative example, Scherhag et al.\ \cite{scherhag2020deep} proposed a D-MAD method based on deep face representations.
In their method, the difference between deep facial features extracted from the document image and the live image is fed into a classifier to distinguish bona-fide images from morphed ones.
In addition, this framework has been extended to address limitations in training data.
For example, IDSwapMAD proposed by Banas et al.\ \cite{banas2026idswapmad} preserves the differential comparison scheme based on AdaFace features and an SVM.
At the same time, it adapts the framework to more privacy-friendly training conditions.

While these methods bypass explicit face reconstruction by directly exploiting recognition feature spaces, their static comparisons remain sensitive to specific models and environmental conditions, often failing to capture morph-specific behaviors.

\subsection{Face De-Morphing-Based D-MAD}
Face de-morphing-based D-MAD detects morphing attacks by applying reverse morphing or separation processes to the document image and the live image.
The decision is then made based on the reconstructed constituent face or the resulting similarity patterns.

A starting point of this line of research was the face de-morphing method proposed by Ferrara et al.\ \cite{ferrara2017face}, which attempted to reconstruct the contributing face by removing the bona-fide component.
Following early reconstruction efforts, recent advancements have leveraged deep generative models; for instance, Long et al.\ \cite{10415238} improved reconstruction accuracy and image quality using a Diffusion Autoencoder.
Furthermore, Hoang et al.\ \cite{hoang2025face} addressed unknown mixing conditions by utilizing a sequence of similarity scores obtained with multiple de-morphing factors.

Overall, face de-morphing-based D-MAD represents a major direction in D-MAD research.
At the same time, its detection performance tends to depend heavily on reconstruction quality and the choice of de-morphing conditions.

\subsection{Hybrid-Based D-MAD}
Hybrid-based D-MAD aims at more robust detection by integrating two types of information.
One is identity discrepancy obtained from the comparison between the document image and the live image.
The other is morphing-related artifacts contained in the document image itself.

Borghi et al.\ \cite{borghi2021double} proposed a double Siamese framework.
It consists of one branch for extracting identity-related features and another branch for detecting morphing artifacts.
In this way, the method jointly exploits differential cues and single-image artifacts.
Similarly, Liu et al.\ \cite{liu2024differential} combined triplet-based metric learning with an artifact extractor.
The former emphasizes discrepancies in the identity feature space.
The latter analyzes morphing-related traces.
Both types of information are then fused for the final decision.

These methods are effective because they combine feature difference-based D-MAD with artifact analysis.
However, their model design tends to be more complex.
Their final performance is also strongly influenced by how and when the different sources of information are integrated.
\begin{figure*}[t]
    \centering
    \includegraphics[width=0.8\linewidth]{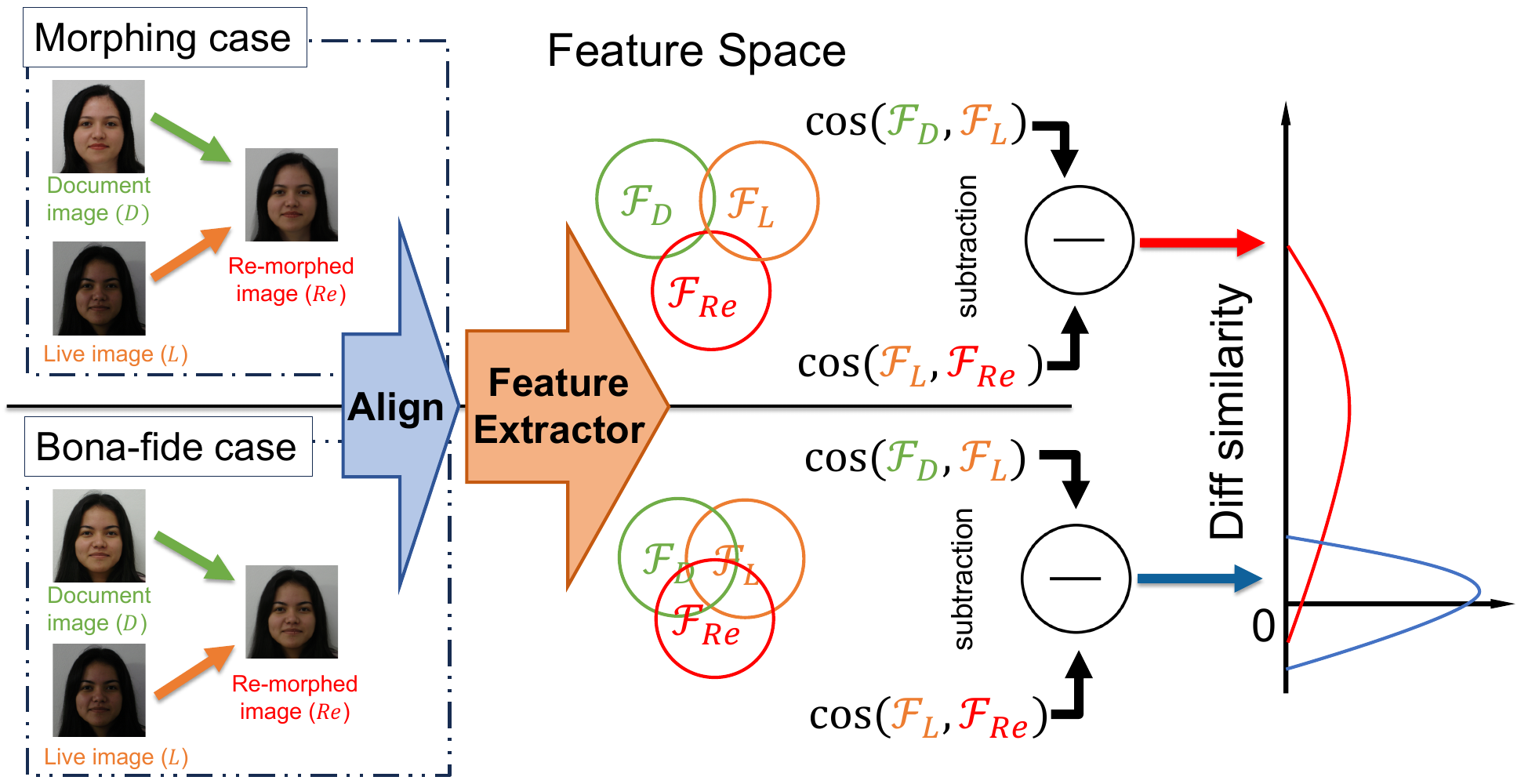}
    \caption{Overview of the proposed method. The proposed framework compares the cosine similarity between the document image and the live image with that between the live image and the re-morphed image, and uses their difference as the detection score.}
    \label{fig:Overview}
\end{figure*}

As outlined above, existing D-MAD research has evolved primarily along three directions: difference-based comparison, constituent-face reconstruction, and the integration of multiple cues.
In contrast, the method investigated in this study is distinguished by incorporating existing morphing techniques into the detection process itself and by using the change in comparison scores before and after re-morphing between the document image and the live image as a discriminative cue.
Accordingly, the proposed method can be positioned as a D-MAD approach that departs from the conventional paradigms of static difference comparison and constituent-face reconstruction.

\section{Proposed Method}

\subsection{Overview}
This study is based on the hypothesis that bona-fide and morphed cases exhibit different patterns of facial feature similarity change in the feature space when a re-morphed image is generated by applying an additional morphing process to a document image and a live image.
The proposed method determines whether a morphing attack is present by using, as a discriminative cue, the difference between the cosine similarity of the document image and the live image and that of the re-morphed image and the live image.
Figure~\ref{fig:Overview} illustrates the overall framework of the proposed method.

\subsection{Re-morphed Image Generation}
In this study, the re-morphing process refers to applying a morphing operation, within the Face Re-morphing framework, to the document image $D$ to be verified and the live image $L$ acquired in a trusted environment.
First, face alignment is applied to $D$ and $L$, and a re-morphed image $Re$ is generated by applying a morphing technique to the aligned images.

This study adopts three generation methods with different underlying principles for the re-morphing process.
MorDIFF is a representation-space morphing technique based on a Diffusion Autoencoder, which generates morphed images by interpolating the latent representations of the input images.
OpenCV is a classical image-level morphing method based on corresponding facial landmarks, which synthesizes images through Delaunay-triangulation-based warping and alpha blending.
StyleGAN2-ADA is a GAN-based method that generates morphed images by manipulating latent representations in the GAN latent space.
By employing these morphing techniques, we examine how the choice of the re-morphing method affects the proposed detection framework.

\subsection{Feature Extraction and Score Computation}
\label{sec:feature_extraction_score}
The proposed method feeds each preprocessed image into an identical facial feature extractor $\mathcal{F}$ to obtain the corresponding feature vectors $\mathcal{F}_D$, $\mathcal{F}_L$, and $\mathcal{F}_{Re}$.
Using these feature vectors, we calculate the baseline cosine similarity $s_{\mathrm{base}}$ between $D$ and $L$, as well as the cosine similarity $s_{\mathrm{re}}$ between $L$ and $Re$, as follows:
\begin{equation}
s_{\mathrm{base}}=\cos(\mathcal{F}_D,\mathcal{F}_L), \quad
s_{\mathrm{re}}=\cos(\mathcal{F}_L,\mathcal{F}_{Re})
\end{equation}

The proposed method uses the difference between these two similarities as the final detection score:
\begin{equation}
\Delta s = s_{\mathrm{re}} - s_{\mathrm{base}}
\end{equation}

This score $\Delta s$ represents the amount of variation introduced by the re-morphing process relative to the original similarity between the document image $D$ and the live image $L$.
% 定性的記述をコメントアウトして3.4にTheoretical Justificationを追加 by tohki
% In a bona-fide case, the document image $D_B$ is composed of the facial characteristics of a single individual.
% Therefore, the re-morphed image $Re_B$ generated from $D_B$ and $L$ is expected to remain relatively close to the original identity relationship in the feature space, resulting in a comparatively small value of $\Delta s$.
% In contrast, in a morphed case, the document image $D_M$ is already a synthetic image containing facial characteristics from multiple individuals.
% As a result, the re-morphed image $Re_M$ generated from $D_M$ and $L$ is expected to exhibit a different feature response from that observed in the bona-fide case.
% Consequently, the distribution of $\Delta s$ for morphed cases is expected to differ from that for bona-fide cases.
Based on this difference, the proposed method performs morphing attack detection by using the cosine similarity change induced by the dynamic operation of re-morphing.

\subsection{Theoretical Justification}
Let $\mathcal{F}_A$ and $\mathcal{F}_C$ denote the facial embeddings of the accomplice and criminal identities, respectively. In the following, we analyze the Criminal scenario where $L=C$ without loss of generality. The same argument applies symmetrically to the Accomplice scenario.

\vspace{1mm}\noindent\textbf{Bona-fide case ($D$ and $L$ belong to the same identity $I$).}
The embeddings $\mathcal{F}_D$ and $\mathcal{F}_L$ correspond to two samples of the same identity and are therefore expected to be close in the embedding space, i.e.,
\[
\mathcal{F}_D \approx \mathcal{F}_L \approx \mathcal{F}_I.
\]
Since modern face recognition models exhibit strong intra-class invariance, the image generated by re-morphing two samples of the same identity is also likely to be recognized as identity $I$, and thus $\mathcal{F}_{Re}$ is expected to remain near the same identity cluster. Therefore, both $\cos(\mathcal{F}_D,\mathcal{F}_L)$ and $\cos(\mathcal{F}_L,\mathcal{F}_{Re})$ reflect intra-class similarity, yielding
\[
\Delta s = \cos(\mathcal{F}_L,\mathcal{F}_{Re}) - \cos(\mathcal{F}_D,\mathcal{F}_L) \approx 0
\]
regardless of moderate intra-class variation.

\vspace{1mm}\noindent\textbf{Morphing case ($D$ is a morph of $A$ and $C$, and $L=C$).}
The embedding $\mathcal{F}_D$ can be approximated as lying in the region between $\mathcal{F}_A$ and $\mathcal{F}_C$. This property has been 
empirically observed in prior morphing attack studies, where morph images are matched against both contributing identities at high similarity scores~\cite{ferrara2014magic, scherhag2020deep}.

Consequently, $\cos(\mathcal{F}_D,\mathcal{F}_L)$ reflects a similarity lower than typical intra-class similarity. When re-morphing is applied between $D$ and $L$, the resulting embedding $\mathcal{F}_{Re}$ tends to move closer to $\mathcal{F}_C$, because $L$ contributes a full identity component while $D$ contains only a partial contribution from $C$. Therefore,
\[
\cos(\mathcal{F}_L,\mathcal{F}_{Re}) > \cos(\mathcal{F}_D,\mathcal{F}_L)
\]
typically holds, yielding a positive $\Delta s$ in many cases, with the magnitude expected to increase as the embedding distance between $\mathcal{F}_A$ and $\mathcal{F}_C$ becomes larger.

The same argument applies symmetrically to the Accomplice scenario where $L=A$. In this case, re-morphing pulls $\mathcal{F}_{Re}$ toward $\mathcal{F}_A$, again yielding a positive $\Delta s$.

\section{Experiments}

\subsection{D-MAD Scenarios}
\label{subsec:Scenarios}

In the D-MAD setting, a document image is compared with a trusted live image to distinguish bona-fide pairs from morphing attack pairs. A bona-fide pair consists of a bona-fide document image and a live image of the same identity, whereas an attack pair consists of a morph document image and a live image of one of the two individuals used to generate the morph. The three conditions differ according to which individual provides the live image in the attack pair.

\noindent\textbf{Accomplice.} The morph document image is compared with a live image of the accomplice, who submits the morph image when applying for the document. This condition mainly represents the enrollment or document-issuance stage, where the morph image must be sufficiently similar to the accomplice to be accepted. Detection performance is evaluated using the bona-fide pairs and the accomplice-side attack pairs.

\noindent\textbf{Criminal.} The morph document image is compared with a live image of the criminal, who subsequently attempts to use the issued document. This condition mainly represents the operational identity-verification stage, where the criminal attempts to be verified against the morph document image. Detection performance is evaluated using the bona-fide pairs and the criminal-side attack pairs.

\noindent\textbf{Both.} The Accomplice and Criminal cases are evaluated together against the bona-fide pairs. This condition therefore provides an overall evaluation that considers both possible users of the morph document, rather than averaging the results obtained separately for the two conditions.

\subsection{Datasets}

In this study, the FRLL-Morphs dataset~\cite{9746477, Sarkar2020} and the FEI Morph dataset~\cite{di2023combining, thomaz2010new} were used for evaluation.

FRLL-Morphs is a family of morphing face datasets constructed from the Face Research Lab London Set~\cite{debruine_jones_2017}.
We evaluated five morphing conditions: AMSL~\cite{Neubert_Tom_Makrushin}, FaceMorpher\footnote{Yao Pang, ``FaceMorpher,'' GitHub repository. Available: \url{https://github.com/yaopang/FaceMorpher}.}, OpenCV\footnote{Satya Mallick, ``Face Morph Using OpenCV -- C++ / Python,'' LearnOpenCV, Mar. 11, 2016. Available: \url{https://learnopencv.com/face-morph-using-opencv-cpp-python/}.}, StyleGAN~\cite{Karras2019stylegan2}, and WebMorph.
All conditions share the same 102 bona-fide comparisons.
The numbers of Accomplice and Criminal comparisons are 2,175 each for AMSL, 1,222 each for FaceMorpher and StyleGAN, and 1,221 each for OpenCV and WebMorph.

The FEI Morph dataset is constructed from the FEI Face Database and consists of Version 1 and Version 2.
Version 1~\cite{raja2020morphing} contains morphs generated by FaceFusion, FaceMorph, and a triangulation variant based on STASM landmarks, while Version 2~\cite{batskos2023visualizing,raja2020morphing} additionally includes FaceMorpher, UBOMorph, a SURYS method, and a method proposed by the University of Twente.
Both versions share 400 bona-fide comparisons.
The numbers of Accomplice and Criminal comparisons are 12,000 each for Version 1 and 28,000 each for Version 2.
Therefore, Version 2 is used for the main analysis, as it subsumes the morphing methods in Version 1 and provides a more comprehensive evaluation set.

The pair information used in our experiments follows the public ACIdA repository~\cite{di2025improving}.
Since the proposed method is training-free, all available pairs were used for evaluation.
This public pair definition avoids manual selection of advantageous test pairs and provides a consistent basis for comparison with reported benchmark results.

\subsection{Re-morphing Parameters}
\label{Re-m Parameters}
For each morphing method used for re-morphing, the experimental settings followed the corresponding official implementation.
For a fair comparison, the two input face images were equally weighted in all methods by setting the interpolation coefficient to $\alpha=0.5$. The sensitivity to $\alpha$ is analyzed in the supplementary material.

For MorDIFF~\cite{MorDIFF, SYNMAD2022}, the official implementation based on a Diffusion Autoencoder was used.
The generation model employed an FFHQ-pretrained configuration at 256 resolution.
As input preprocessing, FFHQ-style face alignment based on 68-point dlib landmarks was applied, and the final aligned image size was set to 256.
Padding was enabled during alignment.
The number of steps for stochastic encoding was set to $T=250$, and the number of steps for final image reconstruction was set to $T=20$.
Under these settings, the re-morphed image was generated from the intermediate representation of the two input face images.

For OpenCV, the same face alignment strategy as in MorDIFF was applied, and the final aligned image size was also set to 256.
Padding was enabled during alignment.
After alignment, auxiliary points along the image boundary were added to the landmarks, Delaunay triangulation was performed, and each triangular region was synthesized using affine transformation followed by linear interpolation between the two images.
Facial landmarks were detected using the dlib 68-point shape predictor, and the number of upsampling steps during face detection was set to 1.

For StyleGAN2-ADA~\cite{Karras2020ada}, faces were first detected using 68-point dlib landmarks, and FFHQ-style face alignment was applied to obtain aligned input images at 1024 resolution.
Each input image was then projected into the $W$ space using the StyleGAN2-ADA projector, and the resulting latent representations were denoted by $w_a$ and $w_b$, respectively.
These latent representations were linearly interpolated as
\begin{equation}
w_{\mathrm{mix}} = (1-\alpha) w_a + \alpha w_b
\end{equation}
and the interpolated latent representation was fed into the generator to synthesize the re-morphed image.
The number of optimization steps for inversion into the $W$ space was set to 1000 for each image.
The rationale for selecting $\alpha=0.5$ is discussed in the supplementary material.

\subsection{Face Feature Extractor}

CVLFace~\cite{9880230, 10655027}, an existing face recognition toolkit, was used as the face feature extractor.
CVLFace provides multiple face recognition backbones and training settings through a unified interface, and in this study, four models trained on WebFace4M, namely IR101, IR50, ViT KPRPE (ViT-k), and ViT,  were used.
For input preprocessing, the Face Alignment App included in CVLFace was used together with its corresponding face alignment model.

\subsection{Evaluation Metrics}

The performance of the proposed method was evaluated using Bona-fide Sample Classification Error Rate (BSCER) and Morph Attack Classification Error Rate (MACER).
BSCER represents the proportion of bona-fide samples that are incorrectly classified as morphs, whereas MACER represents the proportion of morph samples that are incorrectly classified as bona-fide.
These metrics were calculated following the evaluation framework of ISO/IEC 20059 for biometric systems against morphing attacks.

Let the number of bona-fide samples be denoted by $N_{\mathrm{BF}}$, the number of morph samples by $N_{\mathrm{M}}$, and the classification result for each sample by $Res_i$.
Here, $Res_i$ takes the value 1 if the $i$-th sample is classified as a morph, and 0 if it is classified as bona-fide.
Under this definition, BSCER and MACER are given by

\begin{equation}
\mathrm{BSCER}
=
\frac{\sum_{i=1}^{N_{\mathrm{BF}}} Res_i}{N_{\mathrm{BF}}}
\end{equation}

\begin{equation}
\mathrm{MACER}
=
\frac{\sum_{i=1}^{N_{\mathrm{M}}} (1-Res_i)}{N_{\mathrm{M}}}
\end{equation}

In addition, the Equal Error Rate (EER) was calculated as the error rate at the operating point where BSCER and MACER are equal.
Furthermore, the BSCER value obtained when fixing MACER at 1\% was reported as $\mathrm{B}_{0.01}$.
This enables evaluation of the misclassification rate for bona-fide samples under an operating condition where the false acceptance rate for morph samples is constrained to 1\%.

\section{Experimental Results}

In this section, the detection performance of the proposed method on FRLL-Morphs and FEI Morph is evaluated from four perspectives.
(1) Comparison across feature extractors: the effect of the feature extractor on detection performance is analyzed.
Since the proposed method identifies morphing attacks based on the relationships among features extracted from the document image, the live image, and the re-morphed image generated from them, it is important to examine whether consistent tendencies can be observed across different feature extractors.
(2) Comparison across re-morphing methods: the effect of the re-morphing method on detection performance is compared.
In this comparison, the average performance over different feature extractors is used to analyze how the re-morphing methods, namely OpenCV (OCV), MorDIFF (MD), and StyleGAN2-ADA (SG2-A), affect D-MAD performance.
(3) Comparison across detection scores: the proposed score $\Delta s$ is compared with $s_{\mathrm{re}}$, which uses only the similarity between the re-morphed image and the live image.
This comparison verifies the effectiveness of the difference-based score using the similarity change before and after re-morphing.
(4) Comparison with existing methods: the proposed method is compared with existing methods. This comparison is conducted only under conditions where direct comparison with previous studies is possible, and AMSL in FRLL-Morphs and FEI Morph, for which published benchmark results are available, are used to clarify the relative position of the proposed method.

\subsection{Comparison of Feature Extractors}
This subsection compares the performance of different feature extractors on FRLL-Morphs (FRLL-M) and FEI Morph Version 2 (FEI-M2).
The results are shown in Table~\ref{tab:across_feature_extractors_results}.

On FRLL-M, high detection performance was obtained regardless of the feature extractor.
In the OCV setting, performance was largely consistent across feature extractors, with EERs ranging from 0.00099 to 0.0014 across the three conditions, while B$_{0.01}$ was 0 for all extractors and conditions, indicating only minor variation across feature extractors.
MD also showed stable performance, with the mean EER and mean B$_{0.01}$ remaining at most 0.0098.
These results indicate that, on FRLL-M, the proposed method does not strongly depend on a specific feature extractor.

On FEI-M2, the overall error rates were higher than those on FRLL-M, indicating a more challenging condition.
Nevertheless, the overall tendencies were consistent across feature extractors.
The Criminal condition yielded low error rates, whereas the Accomplice condition showed degraded performance.
The standard deviation of EER was relatively small in many conditions, indicating that the basic discriminative tendency of the proposed method was preserved across feature extractors.
On the other hand, some variations were observed in B$_{0.01}$, suggesting that the feature extractor may affect strict operating points such as the low-MACER region.

Overall, the proposed method maintained consistent performance tendencies across different feature extractors, demonstrating that the similarity change caused by re-morphing can be captured in multiple face recognition feature spaces.

\begin{table}[!t]
\centering
\setlength{\tabcolsep}{2pt}
\renewcommand{\arraystretch}{0.95}
\caption{Detection performance using different feature extractors on FRLL-Morphs and FEI Morph Version 2.}
\label{tab:across_feature_extractors_results}
\footnotesize
\begin{tabular}{lcccccc}
\toprule
\multirow{2}{*}{Setting}
& \multicolumn{2}{c}{Accomplice}
& \multicolumn{2}{c}{Criminal}
& \multicolumn{2}{c}{Both} \\
\cmidrule(lr){2-3} \cmidrule(lr){4-5} \cmidrule(lr){6-7}
& EER & B$_{0.01}$
& EER & B$_{0.01}$
& EER & B$_{0.01}$ \\
\midrule
{\footnotesize FRLL-M / MD / IR101} & \enum{1.1}{-2} & \enum{2.0}{-2} & \enum{3.0}{-3} & 0 & \enum{7.6}{-3} & 0 \\
{\footnotesize FRLL-M / MD / IR50} & \enum{8.4}{-3} & 0 & \enum{2.0}{-3} & 0 & \enum{5.2}{-3} & 0 \\
{\footnotesize FRLL-M / MD / ViT-k} & \enum{9.8}{-3} & \enum{9.8}{-3} & \enum{2.3}{-3} & 0 & \enum{7.9}{-3} & 0 \\
{\footnotesize FRLL-M / MD / ViT} & \enum{9.8}{-3} & \enum{9.8}{-3} & \enum{2.1}{-3} & 0 & \enum{7.2}{-3} & 0 \\
{\footnotesize FRLL-M / MD / mean} & \enum{9.7}{-3} & \enum{9.8}{-3} & \enum{2.3}{-3} & 0 & \enum{7.0}{-3} & 0 \\
{\footnotesize FRLL-M / MD / std} & \enum{1.0}{-3} & \enum{8.0}{-3} & \enum{4.4}{-4} & 0 & \enum{1.2}{-3} & 0 \\
\midrule
{\footnotesize FRLL-M / OCV / IR101} & \enum{9.9}{-4} & 0 & \enum{9.9}{-4} & 0 & \enum{9.9}{-4} & 0 \\
{\footnotesize FRLL-M / OCV / IR50} & \enum{9.9}{-4} & 0 & \enum{9.9}{-4} & 0 & \enum{9.9}{-4} & 0 \\
{\footnotesize FRLL-M / OCV / ViT-k} & \enum{9.9}{-4} & 0 & \enum{9.9}{-4} & 0 & \enum{9.9}{-4} & 0 \\
{\footnotesize FRLL-M / OCV / ViT} & \enum{1.4}{-3} & 0 & \enum{9.9}{-4} & 0 & \enum{1.2}{-3} & 0 \\
{\footnotesize FRLL-M / OCV / mean} & \enum{1.1}{-3} & 0 & \enum{9.9}{-4} & 0 & \enum{1.0}{-3} & 0 \\
{\footnotesize FRLL-M / OCV / std} & \enum{2.1}{-4} & 0 & 0 & 0 & \enum{1.1}{-4} & 0 \\
\midrule
{\footnotesize FRLL-M / SG2-A / IR101} & \enum{6.9}{-2} & \enum{2.8}{-1} & \enum{5.5}{-2} & \enum{2.4}{-1} & \enum{5.9}{-2} & \enum{2.5}{-1} \\
{\footnotesize FRLL-M / SG2-A / IR50} & \enum{7.9}{-2} & \enum{2.9}{-1} & \enum{5.2}{-2} & \enum{2.6}{-1} & \enum{6.9}{-2} & \enum{2.9}{-1} \\
{\footnotesize FRLL-M / SG2-A / ViT-k} & \enum{8.0}{-2} & \enum{3.7}{-1} & \enum{5.7}{-2} & \enum{2.7}{-1} & \enum{6.7}{-2} & \enum{3.4}{-1} \\
{\footnotesize FRLL-M / SG2-A / ViT} & \enum{8.8}{-2} & \enum{3.6}{-1} & \enum{6.9}{-2} & \enum{3.1}{-1} & \enum{7.8}{-2} & \enum{3.5}{-1} \\
{\footnotesize FRLL-M / SG2-A / mean} & \enum{7.9}{-2} & \enum{3.3}{-1} & \enum{5.8}{-2} & \enum{2.7}{-1} & \enum{6.8}{-2} & \enum{3.1}{-1} \\
{\footnotesize FRLL-M / SG2-A / std} & \enum{8.0}{-3} & \enum{4.6}{-2} & \enum{7.3}{-3} & \enum{3.2}{-2} & \enum{8.1}{-3} & \enum{4.6}{-2} \\
\midrule
\midrule
{\footnotesize FEI-M2 / MD / IR101} & \enum{9.4}{-2} & \enum{8.1}{-1} & \enum{1.3}{-2} & \enum{1.8}{-2} & \enum{6.6}{-2} & \enum{6.0}{-1} \\
{\footnotesize FEI-M2 / MD / IR50} & \enum{1.2}{-1} & \enum{8.6}{-1} & \enum{1.8}{-2} & \enum{5.0}{-2} & \enum{8.5}{-2} & \enum{6.9}{-1} \\
{\footnotesize FEI-M2 / MD / ViT-k} & \enum{1.1}{-1} & \enum{7.6}{-1} & \enum{6.3}{-3} & \enum{2.5}{-3} & \enum{7.0}{-2} & \enum{5.9}{-1} \\
{\footnotesize FEI-M2 / MD / ViT} & \enum{1.1}{-1} & \enum{7.6}{-1} & \enum{7.2}{-3} & \enum{5.0}{-3} & \enum{6.6}{-2} & \enum{6.0}{-1} \\
{\footnotesize FEI-M2 / MD / mean} & \enum{1.1}{-1} & \enum{8.0}{-1} & \enum{1.1}{-2} & \enum{1.9}{-2} & \enum{7.2}{-2} & \enum{6.2}{-1} \\
{\footnotesize FEI-M2 / MD / std} & \enum{1.3}{-2} & \enum{4.8}{-2} & \enum{5.5}{-3} & \enum{2.2}{-2} & \enum{9.2}{-3} & \enum{5.0}{-2} \\
\midrule
{\footnotesize FEI-M2 / OCV / IR101} & \enum{1.1}{-1} & \enum{5.7}{-1} & \enum{1.1}{-2} & \enum{1.3}{-2} & \enum{8.1}{-2} & \enum{3.5}{-1} \\
{\footnotesize FEI-M2 / OCV / IR50} & \enum{1.4}{-1} & \enum{8.7}{-1} & \enum{2.2}{-2} & \enum{1.3}{-1} & \enum{1.1}{-1} & \enum{7.1}{-1} \\
{\footnotesize FEI-M2 / OCV / ViT-k} & \enum{1.1}{-1} & \enum{3.8}{-1} & \enum{6.0}{-3} & \enum{5.0}{-3} & \enum{7.5}{-2} & \enum{3.1}{-1} \\
{\footnotesize FEI-M2 / OCV / ViT} & \enum{1.1}{-1} & \enum{4.1}{-1} & \enum{8.1}{-3} & \enum{2.5}{-3} & \enum{7.8}{-2} & \enum{3.4}{-1} \\
{\footnotesize FEI-M2 / OCV / mean} & \enum{1.2}{-1} & \enum{5.6}{-1} & \enum{1.2}{-2} & \enum{3.7}{-2} & \enum{8.6}{-2} & \enum{4.3}{-1} \\
{\footnotesize FEI-M2 / OCV / std} & \enum{1.6}{-2} & \enum{2.3}{-1} & \enum{7.0}{-3} & \enum{6.1}{-2} & \enum{1.6}{-2} & \enum{1.9}{-1} \\
\midrule
{\footnotesize FEI-M2 / SG2-A / IR101} & \enum{2.3}{-1} & \enum{8.6}{-1} & \enum{4.8}{-2} & \enum{1.6}{-1} & \enum{1.7}{-1} & \enum{7.7}{-1} \\
{\footnotesize FEI-M2 / SG2-A / IR50} & \enum{2.5}{-1} & \enum{8.7}{-1} & \enum{5.5}{-2} & \enum{1.9}{-1} & \enum{1.7}{-1} & \enum{8.0}{-1} \\
{\footnotesize FEI-M2 / SG2-A / ViT-k} & \enum{2.5}{-1} & \enum{9.1}{-1} & \enum{4.3}{-2} & \enum{1.6}{-1} & \enum{1.6}{-1} & \enum{8.4}{-1} \\
{\footnotesize FEI-M2 / SG2-A / ViT} & \enum{2.5}{-1} & \enum{8.9}{-1} & \enum{3.5}{-2} & \enum{1.9}{-1} & \enum{1.7}{-1} & \enum{8.3}{-1} \\
{\footnotesize FEI-M2 / SG2-A / mean} & \enum{2.4}{-1} & \enum{8.8}{-1} & \enum{4.5}{-2} & \enum{1.7}{-1} & \enum{1.7}{-1} & \enum{8.1}{-1} \\
{\footnotesize FEI-M2 / SG2-A / std} & \enum{8.4}{-3} & \enum{2.1}{-2} & \enum{8.2}{-3} & \enum{1.9}{-2} & \enum{4.6}{-3} & \enum{3.4}{-2} \\

\bottomrule
\end{tabular}
\end{table}

\begin{table}[t!]
\centering
\setlength{\tabcolsep}{2pt}
\renewcommand{\arraystretch}{0.95}
\caption{Average detection performance of different re-morphing methods on FRLL-Morphs and FEI Morph Version 2.}
\label{tab:effect_of_re-morphing}
\footnotesize
\begin{tabular}{lcccccc}
\toprule
\multirow{2}{*}{Setting}
& \multicolumn{2}{c}{Accomplice}
& \multicolumn{2}{c}{Criminal}
& \multicolumn{2}{c}{Both} \\
\cmidrule(lr){2-3} \cmidrule(lr){4-5} \cmidrule(lr){6-7}
& EER & B$_{0.01}$
& EER & B$_{0.01}$
& EER & B$_{0.01}$ \\
\midrule
{\footnotesize FRLL-M / mean / MD   } & \enum{9.7}{-3} & \enum{9.8}{-3} & \enum{2.3}{-3} & 0 & \enum{7.0}{-3} & 0 \\
{\footnotesize FRLL-M / mean / OCV  } & \enum{1.1}{-3} & 0 & \enum{9.9}{-4} & 0 & \enum{1.0}{-3} & 0 \\
{\footnotesize FRLL-M / mean / SG2-A} & \enum{7.9}{-2} & \enum{3.3}{-1} & \enum{5.8}{-2} & \enum{2.7}{-1} & \enum{6.8}{-2} & \enum{3.1}{-1} \\
\midrule
\midrule
{\footnotesize FEI-M2 / mean / MD   } & \enum{1.1}{-1} & \enum{8.0}{-1} & \enum{1.1}{-2} & \enum{1.9}{-2} & \enum{7.2}{-2} & \enum{6.2}{-1} \\
{\footnotesize FEI-M2 / mean / OCV  } & \enum{1.2}{-1} & \enum{5.6}{-1} & \enum{1.2}{-2} & \enum{3.7}{-2} & \enum{8.6}{-2} & \enum{4.3}{-1} \\
{\footnotesize FEI-M2 / mean / SG2-A} & \enum{2.4}{-1} & \enum{8.8}{-1} & \enum{4.5}{-2} & \enum{1.7}{-1} & \enum{1.7}{-1} & \enum{8.1}{-1} \\
\bottomrule
\end{tabular}
\end{table}

\subsection{Comparison of Re-morphing Methods}
\begin{figure}[t!]
    \centering
    \includegraphics[width=\linewidth]{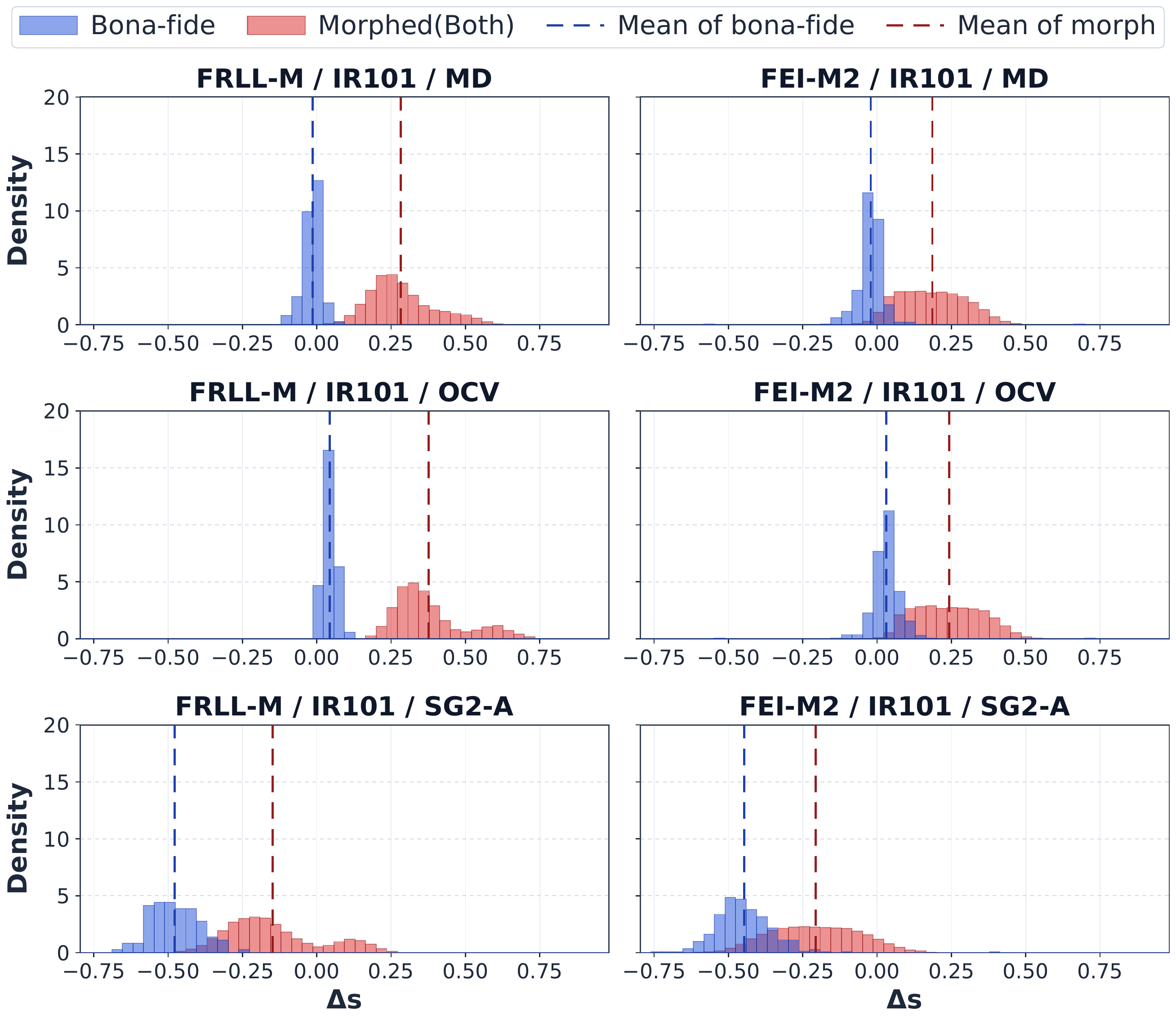}
    \caption{Distributions of $\Delta s$ for bona-fide and morphed samples using IR101 on FRLL-Morphs and FEI Morph Version 2.}
    \label{fig:ir101_delta_s_hist}
\end{figure}

This subsection compares the effect of the morphing method used for re-morphing.
The average values over different feature extractors are reported in Table~\ref{tab:effect_of_re-morphing}.
To further analyze the score distributions, Figure~\ref{fig:ir101_delta_s_hist} shows the $\Delta s$ distributions obtained using IR101 as a representative example.
The results for the other feature extractors are provided in the supplementary material.

On FRLL-M, clear performance differences were observed among the re-morphing methods.
OCV achieved the best performance, with mean EERs of approximately 0.001 and B$_{0.01}$ values of 0 across all conditions, while MD also demonstrated strong detection performance.
In contrast, SG2-A produced substantially higher error rates.
A similar tendency was observed on FEI-M2: MD and OCV performed favorably, whereas SG2-A yielded the highest error rates.
MD achieved the best EER, while OCV outperformed MD in B$_{0.01}$ under the Accomplice and Both conditions.
Across both datasets, the Criminal condition consistently yielded lower error rates than the Accomplice condition, especially for SG2-A.

As shown in Figure~\ref{fig:ir101_delta_s_hist}, for MD and OCV, the $\Delta s$ distribution of bona-fide samples was concentrated around zero, whereas that of morphed samples shifted toward the positive direction.
This tendency is consistent with the assumption of the proposed method.
In particular, the high performance of OCV may be attributed to the stability of landmark-based geometric interpolation, which can reflect the relationship between the input images in a stable manner.

In contrast, SG2-A showed a different distributional tendency.
Since SG2-A requires inversion of the input images into the GAN latent space and subsequent image synthesis from the interpolated latent representation, identity drift may occur due to errors in the inversion and synthesis processes.
As a result, $\mathcal{F}_{Re}$ does not necessarily move closer to $\mathcal{F}_L$, and the expected direction of similarity change may be weakened.
This may contribute to the distributional overlap and the degradation in detection performance.

These results suggest that the proposed method is sensitive not only to artifacts caused by morphing, but also to how consistently the identity relationship is preserved in the feature space through re-morphing.
Therefore, in Face Re-morphing, the choice of the morphing method is important not only for visual quality but also for stable identity changes in the feature space.
A further discussion of the relationship between identity preservation and performance across re-morphing methods is provided in the supplementary material.

\subsection{Comparison of Detection Scores}
This subsection examines the effectiveness of the proposed score $\Delta s$, defined in Section~\ref{sec:feature_extraction_score}, by comparing it with $s_{\mathrm{base}}$ and $s_{\mathrm{re}}$.
The results are shown in Table~\ref{tab:comparison_detection_scores}.

Overall, $\Delta s$ achieved substantially lower EERs than using $s_{\mathrm{re}}$ alone.
On FRLL-M, $\Delta s$ achieved EERs in the range of 0.00099--0.079, whereas those of $s_{\mathrm{re}}$ ranged from 0.044--0.24.
On FEI-M2, the corresponding ranges were 0.011--0.24 and 0.099--0.42, respectively.
These results indicate that the similarity between the re-morphed image and the live image alone is insufficient to separate bona-fide and morphing cases.

Compared with $s_{\mathrm{base}}$, however, the effectiveness of $\Delta s$ depended on the re-morphing method. For OCV, $\Delta s$ outperformed $s_{\mathrm{base}}$ under all three conditions on both datasets. For MD, $\Delta s$ improved the EER under all conditions on FEI-M2, reducing it from 0.15, 0.017, and 0.11 to 0.11, 0.011, and 0.072, respectively, but did not outperform $s_{\mathrm{base}}$ on FRLL-M. In contrast, SG2-A produced higher EERs than $s_{\mathrm{base}}$ under all conditions on both datasets.

These results show that subtracting $s_{\mathrm{base}}$ can reveal similarity changes that are not captured by $s_{\mathrm{re}}$ alone, particularly when MD or OCV is used for re-morphing. However, the degradation observed with SG2-A indicates that the effectiveness of $\Delta s$ depends on the re-morphing method.

% \begin{table}[!t]
% \centering
% \setlength{\tabcolsep}{2pt}
% \renewcommand{\arraystretch}{0.95}
% \caption{Comparison of EERs between the proposed score $\Delta s$ and $s_{\mathrm{re}}$ alone across datasets and re-morphing methods.}
% \label{tab:comparision_Detection_scores}
% \footnotesize
% \begin{tabular}{lcccccc}
% \toprule
% \multirow{2}{*}{Setting}
% & \multicolumn{2}{c}{{\footnotesize Accomplice}}
% & \multicolumn{2}{c}{{\footnotesize Criminal}}
% & \multicolumn{2}{c}{{\footnotesize Both}} \\
% \cmidrule(lr){2-3} \cmidrule(lr){4-5} \cmidrule(lr){6-7}
% & $\Delta s$ & $s_{\mathrm{re}}$
% & $\Delta s$ & $s_{\mathrm{re}}$
% & $\Delta s$ & $s_{\mathrm{re}}$ \\
% \midrule
% {\footnotesize FRLL-M / mean / MD} & \enum{9.7}{-3} & \enum{1.5}{-1} & \enum{2.3}{-3} & \enum{1.1}{-1} & \enum{7.0}{-3} & \enum{1.2}{-1} \\
% {\footnotesize FRLL-M / mean / OCV} & \enum{1.1}{-3} & \enum{7.9}{-2} & \enum{9.9}{-4} & \enum{4.4}{-2} & \enum{1.0}{-3} & \enum{6.2}{-2} \\
% {\footnotesize FRLL-M / mean / SG2-A} & \enum{7.9}{-2} & \enum{2.4}{-1} & \enum{5.8}{-2} & \enum{2.0}{-1} & \enum{6.8}{-2} & \enum{2.2}{-1} \\
% \midrule
% \midrule
% {\footnotesize FEI-M2 / mean / MD} & \enum{1.1}{-1} & \enum{3.4}{-1} & \enum{1.1}{-2} & \enum{1.3}{-1} & \enum{7.2}{-2} & \enum{2.5}{-1} \\
% {\footnotesize FEI-M2 / mean / OCV} & \enum{1.2}{-1} & \enum{3.1}{-1} & \enum{1.2}{-2} & \enum{9.9}{-2} & \enum{8.6}{-2} & \enum{2.3}{-1} \\
% {\footnotesize FEI-M2 / mean / SG2-A} & \enum{2.4}{-1} & \enum{4.2}{-1} & \enum{4.5}{-2} & \enum{2.4}{-1} & \enum{1.7}{-1} & \enum{3.4}{-1} \\
% \bottomrule
% \end{tabular}
% \end{table}

\begin{table}[!t]
\centering
\setlength{\tabcolsep}{5pt}
\renewcommand{\arraystretch}{0.95}
\caption{Comparison of EERs for $s_{\mathrm{base}}$, $\Delta s$, and $s_{\mathrm{re}}$ across datasets and re-morphing methods. Since $s_{\mathrm{base}}$ is independent of the re-morphing method, it is reported once per dataset.}
\label{tab:comparison_detection_scores}
\footnotesize

\begin{tabular}{llccc}
\toprule
Setting
& Score
& Accomplice
& Criminal
& Both \\
\midrule

\multirow{1}{*}{FRLL-M / mean / --}
& $s_{\mathrm{base}}$
& \enum{5.3}{-3}
& \enum{1.2}{-3}
& \enum{4.3}{-3} \\
\cmidrule(lr){1-5}

\multirow{2}{*}{FRLL-M / mean / MD}
& $\Delta s$
& \enum{9.7}{-3}
& \enum{2.3}{-3}
& \enum{7.0}{-3} \\
& $s_{\mathrm{re}}$
& \enum{1.5}{-1}
& \enum{1.1}{-1}
& \enum{1.2}{-1} \\
\cmidrule(lr){1-5}

\multirow{2}{*}{FRLL-M / mean / OCV}
& $\Delta s$
& \enum{1.1}{-3}
& \enum{9.9}{-4}
& \enum{1.0}{-3} \\
& $s_{\mathrm{re}}$
& \enum{7.9}{-2}
& \enum{4.4}{-2}
& \enum{6.2}{-2} \\
\cmidrule(lr){1-5}

\multirow{2}{*}{FRLL-M / mean / SG2-A}
& $\Delta s$
& \enum{7.9}{-2}
& \enum{5.8}{-2}
& \enum{6.8}{-2} \\
& $s_{\mathrm{re}}$
& \enum{2.4}{-1}
& \enum{2.0}{-1}
& \enum{2.2}{-1} \\

\midrule
\midrule

\multirow{1}{*}{FEI-M2 / mean / --}
& $s_{\mathrm{base}}$
& \enum{1.5}{-1}
& \enum{1.7}{-2}
& \enum{1.1}{-1} \\
\cmidrule(lr){1-5}

\multirow{2}{*}{FEI-M2 / mean / MD}
& $\Delta s$
& \enum{1.1}{-1}
& \enum{1.1}{-2}
& \enum{7.2}{-2} \\
& $s_{\mathrm{re}}$
& \enum{3.4}{-1}
& \enum{1.3}{-1}
& \enum{2.5}{-1} \\
\cmidrule(lr){1-5}

\multirow{2}{*}{FEI-M2 / mean / OCV}
& $\Delta s$
& \enum{1.2}{-1}
& \enum{1.2}{-2}
& \enum{8.6}{-2} \\
& $s_{\mathrm{re}}$
& \enum{3.1}{-1}
& \enum{9.9}{-2}
& \enum{2.3}{-1} \\
\cmidrule(lr){1-5}

\multirow{2}{*}{FEI-M2 / mean / SG2-A}
& $\Delta s$
& \enum{2.4}{-1}
& \enum{4.5}{-2}
& \enum{1.7}{-1} \\
& $s_{\mathrm{re}}$
& \enum{4.2}{-1}
& \enum{2.4}{-1}
& \enum{3.4}{-1} \\

\bottomrule
\end{tabular}
\end{table}
\begin{table}[t]
\centering
\setlength{\tabcolsep}{2pt}
\renewcommand{\arraystretch}{0.95}
\caption{Comparison between the proposed method and existing methods on the AMSL dataset. Bold and underlined values indicate the best and second-best results, respectively.}
\label{tab:amsl_comparison}
\footnotesize
\begin{tabular}{lcccccc}
\toprule
\multirow{2}{*}{Method}
& \multicolumn{2}{c}{Accomplice}
& \multicolumn{2}{c}{Criminal}
& \multicolumn{2}{c}{Both} \\
\cmidrule(lr){2-3} \cmidrule(lr){4-5} \cmidrule(lr){6-7}
& EER & B$_{0.01}$
& EER & B$_{0.01}$
& EER & B$_{0.01}$ \\
\midrule
\cite{scherhag2018towards} & .059 & .833 & .089 & .980 & .075 & .966 \\
\cite{scherhag2018morph}   & .069 & .598 & .059 & .667 & .069 & .662 \\
\cite{soleymani2021mutual} & -    & -    & -    & .031 & -    & -    \\
\cite{scherhag2020deep}    & .003 & \underline{.020} & \textbf{.000} & \textbf{.000} & \underline{.001} & .020 \\
\cite{borghi2021double}   & \underline{.001} & \underline{.020} & \underline{.001} & .059 & \underline{.001} & .059 \\
\midrule
Ours &      &      &      &      &      &       \\
{\footnotesize AMSL / mean / MD}   & .014 & .032 & \underline{.001} & \textbf{.000} & .009 & \underline{.005} \\
{\footnotesize AMSL / mean / OCV}     & \textbf{.000} & \textbf{.000} & \textbf{.000} & \textbf{.000} & \textbf{.000} & \textbf{.000} \\
{\footnotesize AMSL / mean / SG2-A}  & .104 & .365 & .023 & \underline{.056} & .070 & .292 \\
\bottomrule
\end{tabular}
\end{table}

\begin{table}[t]
\centering
\setlength{\tabcolsep}{2pt}
\renewcommand{\arraystretch}{0.95}
\caption{Comparison between the proposed method and existing methods on FEI Morph Version 1. Bold and underlined values indicate the best and second-best results, respectively.}
\label{tab:fei_comparison}
\footnotesize
\begin{tabular}{lcccccc}
\toprule
\multirow{2}{*}{Method}
& \multicolumn{2}{c}{Accomplice}
& \multicolumn{2}{c}{Criminal}
& \multicolumn{2}{c}{Both} \\
\cmidrule(lr){2-3} \cmidrule(lr){4-5} \cmidrule(lr){6-7}
& EER & B$_{0.01}$
& EER & B$_{0.01}$
& EER & B$_{0.01}$ \\
\midrule
\cite{ferrara2017face} & .160 & .618 & .027 & .050 & .111 & .522 \\
\cite{scherhag2018towards} & .200 & .623 & .185 & .697 & .192 & .628 \\
\cite{scherhag2020deep} & .178 & .770 & .068 & .295 & .128 & .690 \\
\cite{borghi2021double} & .153 & .563 & .061 & .370 & .115 & .515 \\
\cite{di2023combining} & .125 & .468 & .125 & .440 & .125 & .445 \\
\cite{di2024dealing} & .102 & \underline{.333} & \underline{.023} & \underline{.027} & .070 & \underline{.280} \\
\cite{liu2024differential} & \textbf{.078} & \textbf{.283} & .037 & .115 & \textbf{.059} & \textbf{.230} \\
\midrule
Ours              &      &      &      &      &      &                        \\
{\footnotesize FEI-M1 / mean / MD} & \underline{.099} & .724 & \textbf{.010} & \textbf{.015} & \underline{.066} & .538 \\
{\footnotesize FEI-M1 / mean / OCV} & .105 & .465 & \textbf{.010} & .034 & .079 & .371 \\
{\footnotesize FEI-M1 / mean / SG2-A} & .242 & .882 & .039 & .128 & .167 & .811 \\
\bottomrule
\end{tabular}
\end{table}

\subsection{Comparison with Existing Methods}

This subsection compares the proposed method with existing methods on AMSL and FEI Morph Version 1 (FEI-M1), where direct comparison with previous studies is possible.
The existing results on AMSL and FEI-M1 were taken from Borghi et al.~\cite{borghi2021double} and Liu et al.~\cite{liu2024differential}, respectively.
As described in Section~\ref{subsec:Scenarios}, our evaluation uses publicly defined pair information and does not involve manual selection of test pairs that could favor the proposed method.
However, since the reported results of existing methods may have been obtained under pair settings and experimental protocols that are not fully identical to ours, the comparison should be interpreted as a reference comparison based on published benchmark results.
For consistency with prior studies, the proposed method is reported using the same decimal precision. The results are shown in Tables~\ref{tab:amsl_comparison} and~\ref{tab:fei_comparison}.

On AMSL, the OCV setting achieved the best performance under the Accomplice, Criminal, and Both conditions, with EER and B$_{0.01}$ values of 0.000.
This indicates that no errors were observed under this evaluation protocol. We interpret this result as near-perfect performance on the AMSL benchmark rather than evidence of universal separability. 
This near-perfect result is consistent with the general distributional tendency observed in Figure~\ref{fig:ir101_delta_s_hist}  for OCV-based re-morphing on FRLL-Morphs, where bona-fide and morphed samples show clear separation across morphing conditions. Note that Figure~\ref{fig:ir101_delta_s_hist} aggregates all FRLL-Morphs morphing conditions and is therefore not specific to AMSL.
Since all pairs follow the public ACIdA protocol and no manual pair selection was performed, this result reflects the behavior of the proposed score under this benchmark setting. MD also showed competitive performance, whereas SG2-A yielded higher error rates.

On FEI-M1, the proposed method was particularly effective under the Criminal condition.
Both MD and OCV achieved an EER of 0.010, outperforming the existing methods, and MD achieved the best B$_{0.01}$ value of 0.015.
However, under the Accomplice and Both conditions, the best existing method~\cite{liu2024differential} outperformed the proposed variants.
Thus, the overall performance on FEI-M1 is mainly limited by the Accomplice condition.

This difference may be partly related to the morphing factor.
AMSL uses a fixed morphing factor of 0.5, where the identity components of the accomplice and criminal are relatively balanced.
In contrast, FEI Morph uses morphing factors of 0.3 and 0.5, where the factor denotes the criminal's contribution.
At a factor of 0.3, the accomplice contribution becomes dominant.
Thus, under the Accomplice condition, the morph image can be closer to the live accomplice image, resulting in a higher $s_{\mathrm{base}}$ and a smaller change after re-morphing.
This may increase the overlap between the $\Delta s$ distributions of bona-fide and morphing cases.

Overall, the proposed method outperformed existing methods on AMSL and showed favorable results under the Criminal condition on FEI-M1.
Improving discrimination under the Accomplice condition remains future work.

\section{Conclusion}

This paper proposed Face Re-morphing, a D-MAD method that uses the cosine similarity change caused by an additional morphing operation between the document image and the live image.
The proposed method generates a re-morphed image and uses the difference between the document--live and live--re-morphed similarities as the detection score, thereby incorporating the morphing operation itself into the detection process.

Experiments on FRLL-Morphs and FEI Morph showed that the proposed cue is effective across different morphing conditions, re-morphing methods, and face feature extractors.
The proposed method achieved particularly strong performance on AMSL and low error rates under the Criminal condition on FEI Morph, demonstrating that re-morphing-induced similarity change is a useful cue for D-MAD.

The results also indicate that the method depends on the re-morphing strategy and its interaction with the face feature space.
This study focused on digital-image evaluation, and future work will examine robustness under print--scan and other operational acquisition processes.

\section*{Acknowledgment}
This work was supported in part by JSPS KAKENHI JP23H00463, JP23K28085, and JST Moonshot R\&D Grant Number JPMJMS2215.

{\small
\bibliographystyle{ieee}
\bibliography{egbib}
}

\end{document}